\documentclass[10pt,twocolumn,letterpaper]{article}

\usepackage{main}              
\usepackage{times}  
\usepackage{helvet}  
\usepackage{courier}  
\usepackage[hyphens]{url}  
\usepackage{graphicx} 
\usepackage{natbib}  
\usepackage{array}
\usepackage{booktabs}
\usepackage{bibentry}
\usepackage{makecell}
\usepackage{multirow}
\usepackage{algorithm}
\usepackage{algorithmic}
\usepackage{enumitem}
\usepackage{color}
\usepackage{newfloat}
\usepackage{listings}
\usepackage{lipsum}

\definecolor{cvprblue}{rgb}{0.21,0.49,0.74}
\usepackage[pagebackref,breaklinks,colorlinks,allcolors=cvprblue]{hyperref}

\title{\includegraphics[height=2.5em]{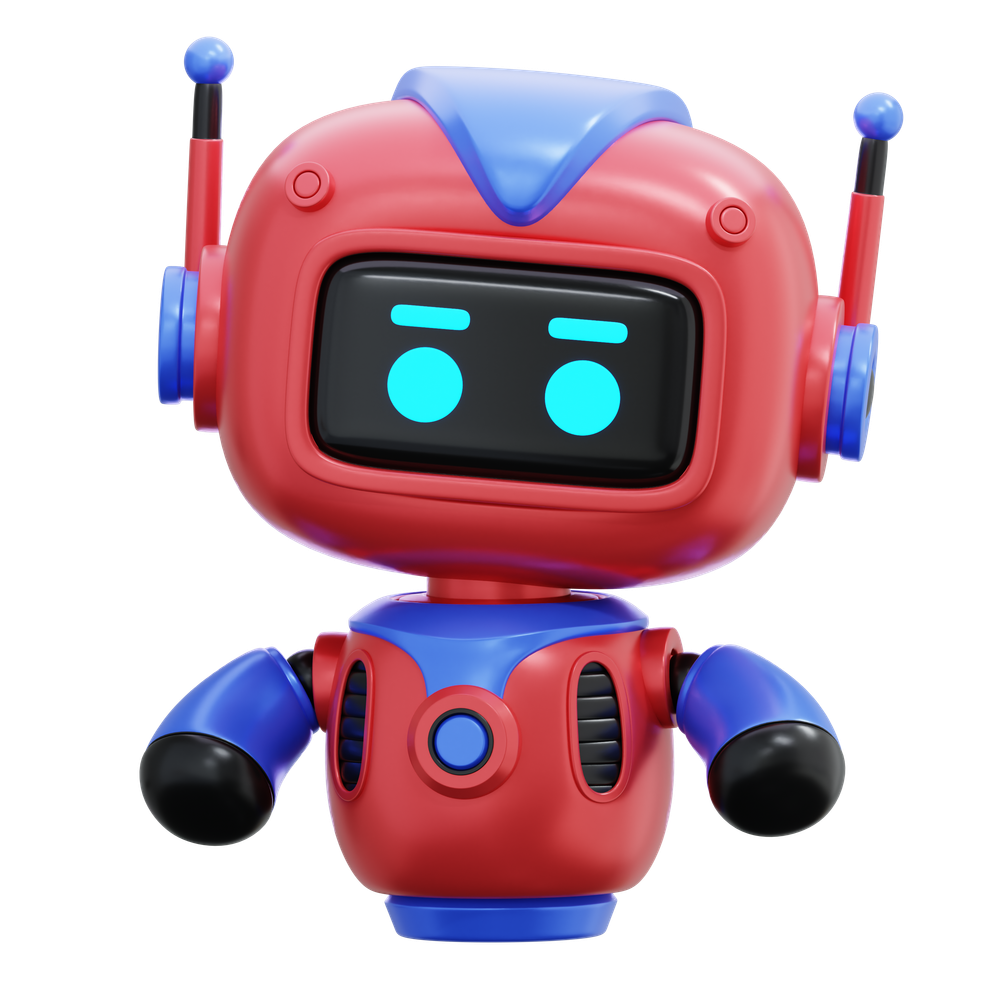}Follow-Your-Instruction: A Comprehensive MLLM Agent for World Data Synthesis}

\author{Kunyu Feng$^{1*}$,  Yue Ma$^{2*}$,  Xinhua Zhang$^{3*}$ \\
Boshi Liu$^{4}$, Yikuang Yuluo$^{5}$, Yinhan Zhang$^{1}$, Runtao Liu$^{2}$, Hongyu Liu$^{2}$, Zhiyuan Qin$^{6}$, Shanhui Mo,\\ Qifeng Chen$^{2\dag}$, Zeyu Wang$^{1,2\dag}$\\
$^1$HKUST(GZ), $^2$HKUST, $^3$Tsinghua Univerisity, $^4$Peking University, \\ $^5$Chongqing University, $^6$Beijing Innovation Center of Humanoid Robotics
}

\begin{document}

\maketitle
\newcommand\blfootnote[1]{%
\begingroup
\renewcommand\thefootnote{}\footnote{#1}%
\addtocounter{footnote}{-1}%
\endgroup
}
\blfootnote{Preprint, Under Review.}
\blfootnote{$^*$ Equal Contribution. $^\dag$ Corresponding Authors.}

\begin{figure*}
    \centering
    \includegraphics[width=0.95\linewidth]{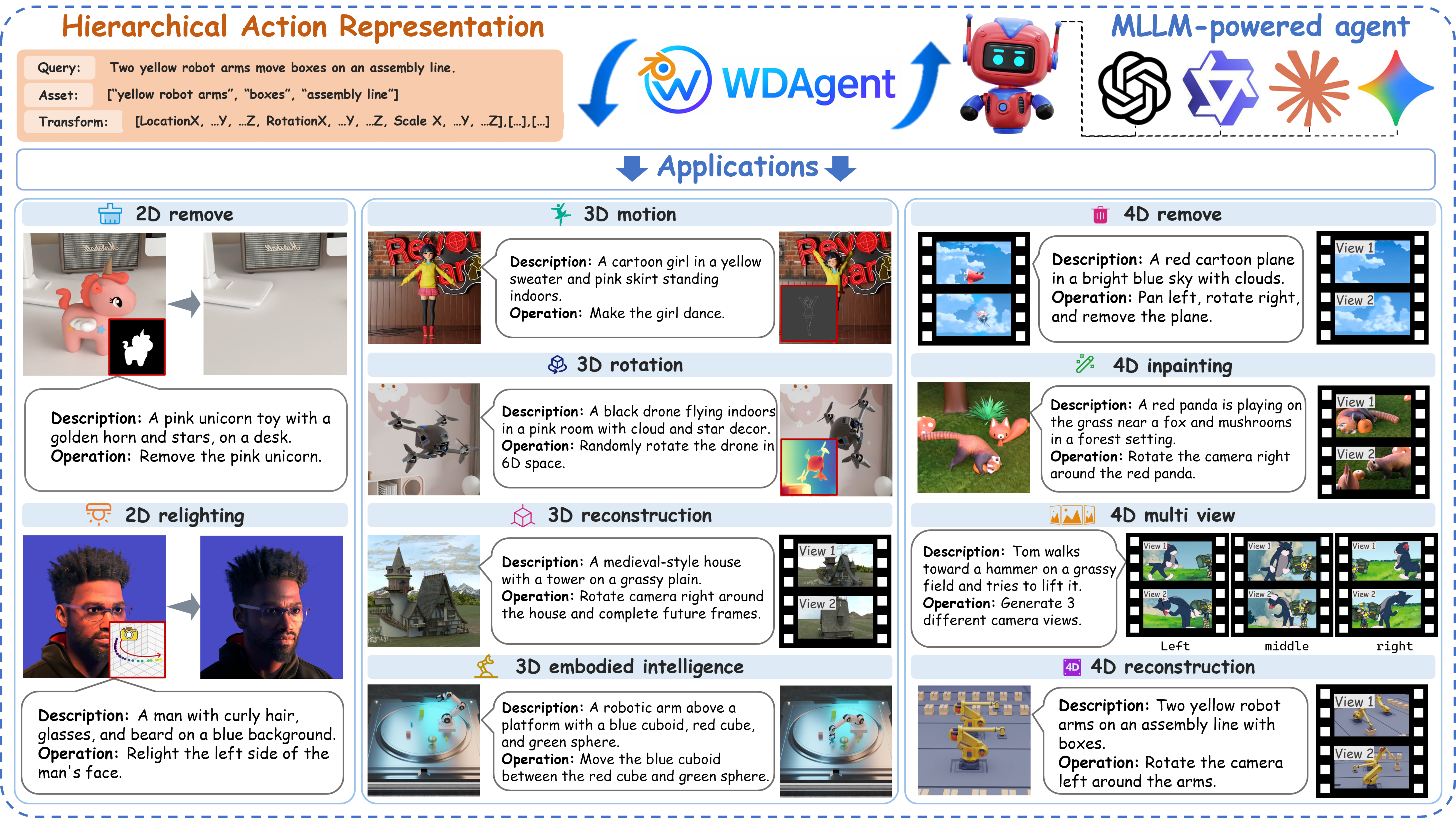}
    \captionof{figure}{\textbf{Overview of Follow-Your-Instruction.}  We introduce Follow-Your-Instruction, an advanced MLLM-driven agent framework that synthesizes high-quality world data across 2D, 3D, and 4D levels, benefiting various downstream applications.
    } 
    \label{fig:Overview}
\end{figure*}
\begin{abstract}
With the growing demands of AI-generated content (AIGC), the need for high-quality, diverse, and scalable data has become increasingly crucial. However, collecting large-scale real-world data remains costly and time-consuming, hindering the development of downstream applications. While some works attempt to collect task-specific data via a rendering process, most approaches still rely on manual scene construction, limiting their scalability and accuracy. To address these challenges, we propose \textbf{Follow-Your-Instruction}, a Multimodal Large Language Model (MLLM)-driven framework for automatically synthesizing high-quality 2D, 3D, and 4D data. Our \textbf{Follow-Your-Instruction} first collects assets and their associated descriptions through multimodal inputs using the \textit{MLLM-Collector}. Then it constructs 3D layouts, and leverages Vision-Language Models (VLMs) for semantic refinement through multi-view scenes with the \textit{MLLM-Generator} and \textit{MLLM-Optimizer}, respectively. Finally, it uses \textit{MLLM-Planner} to generate temporally coherent future frames. We evaluate the quality of the generated data through comprehensive experiments on the 2D, 3D, and 4D generative tasks. The results show that our synthetic data significantly boosts the performance of existing baseline models, demonstrating Follow-Your-Instruction's potential as a scalable and effective data engine for generative intelligence. 
\end{abstract}

\section{Introduction}
\label{sec:intro}

AI-generated content (AIGC) targets to generate creative and realistic content using a generative model. It has been widely applied in the film industry, augmented reality, automated advertising, and creating content for social media.
Recent achievements in the foundation models, such as the diffusion models~\cite{rombach2022high, peebles2023scalable}, Multimodal Large Language Models (MLLMs)~\cite{GPT-4o, Claude-3.5-Sonnet} have significantly enhanced the quality and flexibility of generated content. 
As the data-driving model, these models acquire strong prior knowledge from the large-scale training datasets, which enables them to easily handle challenging tasks, including multi-modal understanding~\cite{qu2025tokenflow, xieshow,ma2022visual,xiao2024bridging} and generation~\cite{flux2024, wan2025wan,ma2024followyouremoji,ma2025followyourmotion,ma2024followpose,chen2024follow,zhu2025multibooth,liao2024freehand}, visual editing~\cite{liu2024video,feng2025dit4edit,zhuinstantswap,yan2025eedit,wang2024cove,ma2025magicstick,wangtaming}, animation~\cite{ma2025followyourclick,zhang2025magiccolor,xue2024follow,xuetowards}, and embodied robots~\cite{liuagentbench,yuemllm}. 

However, as AIGC applications continue to evolve toward more complex and fine-grained scenarios, the demand for high-quality, task-specific data has significantly increased. While most open-source foundation models are trained on large-scale but generic datasets such as LAION-400M~\cite{schuhmann2021laion-400m} and WebVid-10M~\cite{Bain21WebVid-10M}, these datasets typically lack the task-specific annotations required for fine-grained applications. For instance, tasks such as object removal demand accurate background masks, while 4D generation requires accurate camera trajectories. This absence of precise supervision signals often limits the direct applicability of these datasets to specialized generative tasks~\cite{yoshihashi2024exploring}. 

Currently, there are some early works~\cite{rawal2023synthetic, lips2024evaluating} to build task-specific datasets, particularly through rendering pipelines. Rendering engines such as Blender~\cite{blender} enable fine-grained control over object layout, lighting conditions, and physical interactions, making them suitable for constructing datasets tailored to specific AIGC tasks. These synthetic datasets are often used to fine-tune powerful foundation models for improved performance in downstream applications. However, manually designing and curating such datasets remains a significant bottleneck, as it requires substantial human effort, domain expertise, and often struggles to balance realism, accuracy, and scalability~\cite{mumuni2024survey,ma2025controllable}.

To address these limitations, we introduce our \textbf{Follow-Your-Instruction}, an efficient MLLM-based data-synthetic agent framework designed to generate realistic and diverse world data for a wide range of AIGC tasks. More importantly, to the best of our knowledge, our benchmark is the first data generation system that supports both 2D, 3D, and 4D generative tasks. As illustrated in Fig.~\ref{fig:Overview}, our agent encompasses seven representative applications, including 2D object removal, 3D restoration, inpainting, and 4D multi-view generation. In detail, by leveraging the extensive real-world understanding and interactive capabilities of MLLMs, we incorporate strong MLLMs into our agent and introduce four key components, including \textit{MLLM-Collector}, \textit{MLLM-Generator}, \textit{MLLM-Optimizer}, and \textit{MLLM-Planner}, to assist in the design and validation of our benchmark. 

We mainly evaluate the performance of our proposed \textbf{Follow-Your-Instruction} in two scopes: 
(1) Evaluating the MLLM-Driven synthetic data quality: To benchmark the ability of MLLM-Driven synthesis, we perform the experiments on 8 MLLMs, including both commercial tools and research methods, on 4 metrics.
(2) Evaluation on several Downstream Applications: To further evaluate the effectiveness of synthetic data, we finetune 3 various downstream tasks using our synthetic dataset, such as the 2D object removal task, 3D reconstruction, and 4D video generation. The results show substantial improvements in task-specific performance, highlighting the practical benefits of our framework.

In summary, our contributions are as follows:
\begin{itemize}[noitemsep,topsep=0pt]
    \item We propose an efficient MLLM-based data-synthetic agent framework, Follow-Your-Instruction, which synthesizes realistic world data for diverse AIGC tasks. 

    \item To achieve high-quality and efficient data generation, we introduce a comprehensive benchmark to evaluate MLLM-based data-synthetic agents at 2D, 3D, and 4D levels. Additionally, we develop various forms of MLLM-assisted data generation, including in-context and long-term guidance.
    \item To validate the practical performance of our proposed agent, we finetune 3 recent baseline models across representative 2D, 3D, and 4D tasks. Experimental results show that incorporating our data significantly enhances the performance of these models on their respective downstream applications.
\end{itemize}
\section{Related Work}
\label{sec:formatting}

\noindent\textbf{Multi-modal Large Language Models.}
Multi-modal Large Language Models (MLLMs) are advancing by integrating text, vision, and 3D modalities. In content restoration, RestoreAgent~\cite{chen2024restoreagent} shows strong performance in 2D tasks, while RL-Restore~\cite{yu2018crafting} focuses on progressive recovery for blur and noise. Clarity ChatGPT~\cite{wei2023clarity} combines conversation with restoration but remains narrow in scope. For spatial modeling, Text2World~\cite{hu2025text2world} and Spatial-MLLM~\cite{wu2025spatial} focus on symbolic structure generation and dual-encoder-based reasoning, respectively. VSI-Bench~\cite{yang2025thinking} benchmarks spatial reasoning tasks like counting and navigation. In embodied interaction, models like GEA show strong performance on VisualAgentBench~\cite{liuvisualagentbench}, while EmbodiedBench~\cite{yangembodiedbench} reveals limitations in long-term planning for models such as GPT-4V. Despite these advances, challenges like the lack of unified multimodal evaluation and training data remain.

\begin{figure*}[t]
    \centering
    \includegraphics[width=0.98\textwidth]{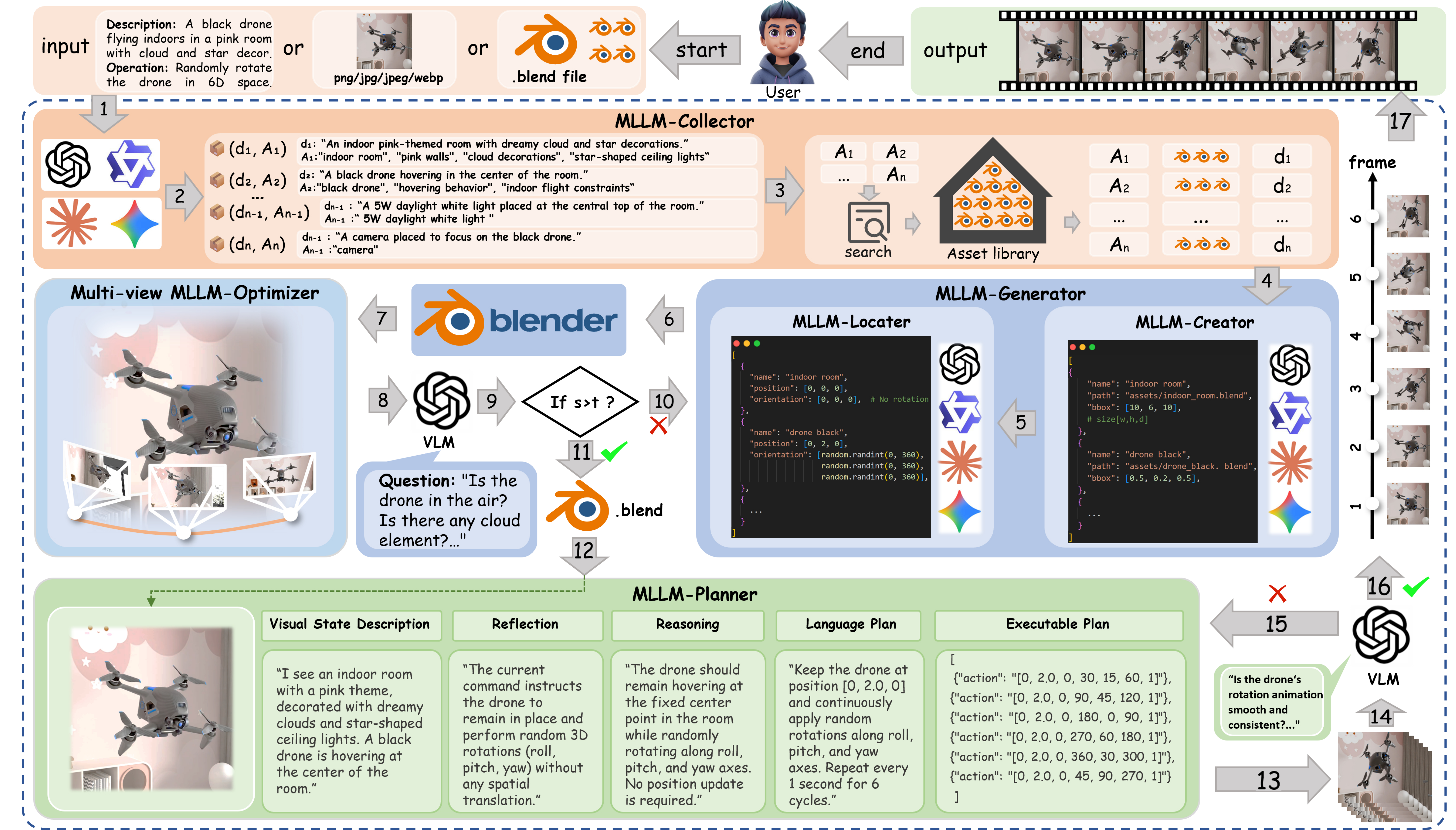}

    \caption{\textbf{The Pipeline of Follow-Your-Instruction.} Given multimodal inputs, Follow-Your-Instruction first collects the assets and their descriptions via \textit{MLLM-Collector}. Then the \textit{MLLM-Generator} creates the 3D layout scene and optimizes the scene via multi-view \textit{MLLM-Optimizer} with a powerful VLM. Based on the scene, \textit{MLLM-Planner} formulates a clear plan to generate the high-quality output video.
    } 
    \label{fig:pipeline}
\end{figure*}

\noindent\textbf{Diffusion-based Generative Applications.}
Diffusion models are widely applied in generative tasks across 2D, 3D, and 4D domains. For 2D tasks like object removal and relighting, prior works~\cite{jiang2025smarteraser, mei2025lux, li2025recap,lun2025towards} rely on manually curated datasets and segmentation pipelines. In 3D, LiDAR Diffusion Models~\cite{ran2024towards} reconstruct depth/point clouds using dedicated datasets, while MV-Adapter~\cite{huang2024mv} ensures multi-view consistency via a plug-and-play module. For 4D, methods like ReCamMaster~\cite{bai2025recammaster} and TrajectoryCrafter~\cite{yu2025trajectorycrafter} leverage 3D structures to ensure cross-camera consistency in video generation. Follow-Your-Creation~\cite{ma2025follow} explores 4D video editing frameworks. However, these methods require large-scale datasets, which are costly to obtain. Our Follow-Your-Instruction addresses this by using MLLMs to generate high-quality synthetic data, reducing real-world data dependency and enhancing adaptability.
\section{Method}

In this section, we introduce our proposed agent, a MLLM-based, comprehensive benchmark across 2D, 3D, and 4D levels. The pipeline of our agent is shown in Fig.~\ref{fig:pipeline}, which is built upon the advanced multimodal Large Language Models (e.g., GPT-4o~\cite{GPT-4o}, QWEN3~\cite{yang2025qwen3technicalreport}). From Sec.~\ref{sec:assets construction} to Sec.~\ref{sec:frame prediction}, we present the details of our proposed agent.

\subsection{Assets Collection via Multimodal Inputs}
\label{sec:assets construction}
Given a conditional input (e.g., an image $I$, a text $T$, or an action $A$), we aim to create high-quality scenes and keep both spatial and temporal consistency. Recent work like SceneCraft~\cite{hu2024scenecraft} designs the LLM-decomposer to generate a list of assets and the description of each sub-scene, then this information can be used for scene generation. Although this method provides a structured decomposition pipeline, it is inherently constrained by the limitations of the input. In particular, complex visual concepts and styles are often difficult to fully express through language alone, which in turn restricts the user’s ability to customize the generative scene.

Our proposed agent incorporates a multimodal asset retrieval mechanism that leverages MLLMs to incorporate both textual and visual information during the asset discovery process. As shown in Fig.~\ref{fig:pipeline}, in addition to prompting with natural language, users can supply specific modalities such as reference images or specific objects. These inputs offer greater flexibility and control, allowing users to specify creative intent in diverse ways. Specifically, we first utilize the MLLMs to transform the inputs into the assets list: 
\begin{equation}
    (d_1, A_1), ... , (d_k, A_k) \gets \textit{MLLM-Collector}(I),
\end{equation}
where the $I$ is inputs, $d_i$ is the sub-scene description, $A_i$ is the list of assets. For text input, we follow the settings from the SceneCraft~\cite{hu2024scenecraft} and apply a top-$k$ retrieval strategy based on relevance scores from our asset repository, selecting the most semantically matched assets for further composition. As for visual input, our agent skips the selection process and directly integrates the assets into the scene construction pipeline. This design not only improves the grounding and fidelity of the generated scenes but also greatly enhances controllability, allowing for more precise and expressive scene authoring. Therefore, our framework significantly enhances the customizable and user-friendly controllability of the scene creation process, making it more suitable for real-world content generation tasks.
\subsection{Global Scenes Construction and Optimization}
\noindent\textbf{3D Layout Generation.} After acquiring the assets from the multimodal inputs, the next step is to integrate all assets and create the entire scene. This process is managed by our \textit{MLLM-Generator}, which performs object generation, spatial placement, and coordinate transformation. Specifically, given an asset $A_i$ and its description $D_i$, it first generates the objects and bounding box in 3D space: 
\begin{equation}
    {B}_i = MLLM-Creator({A}_i, {D}_i),
\end{equation}
where ${B}_i \in {R}^{6}$ represents the 3D bounding box parameters of object $i$ (e.g., center position, width, height, depth). These parameters are adapted to ensure consistency between appearance and semantics.

Then each object is assigned a 3D location ${p}_i^{{world}}$ through the \textit{MLLM-Locator}, which is in the global layout matrix  $\textit{L}_{\textit{world}}$. In detail, there are two placement strategies, one of which is \textbf{\textit{Human-Instruction-Guided Placement}}, which places the object through the specific location in the input instruction as follows:
\begin{equation}
    {p}_i^{{\textit{world}}} = {p}_i^{{\textit{target}}},
\end{equation}

The other is the \textbf{\textit{default strategy}} that the object is placed by aligning its bottom-center point ${c}_i$ to a suitable unoccupied region in the world matrix:
\begin{equation}
    {p}_i^{{\textit{world}}} = {\textit{FindFreeRegion}}({L}_{{world}}, {c}_i),
\end{equation}

Then, we apply a transformation matrix to embed the object into the global layout:
\begin{equation}
L_{\textit{world}}[i] = \textit{ComposeTransform}(p_i^{\textit{world}}, \mathbf{R}_i, \mathbf{s}_i),
\end{equation}
where $\mathbf{R}_i$ and $\mathbf{s}_i$ are the estimated rotation and scale matrices respectively.

Finally, the \textit{MLLM-Locator} projects the 3D layout into the 2D image plane by the given intrinsic matrix $\mathbf{K}$ and extrinsic pose $\mathbf{E}$ of calibrated camera:
\begin{equation}
    {u}_i = \pi(\mathbf{K}, \mathbf{E}, {p}_i^{{\textit{world}}}),
\end{equation}
where $\pi(\cdot)$ denotes the perspective projection function, and $\mathbf{u}_i \in {R}^2$ is the image coordinate of object $i$.

\begin{figure}[t!]
    \centering
    \includegraphics[width=0.88\columnwidth]{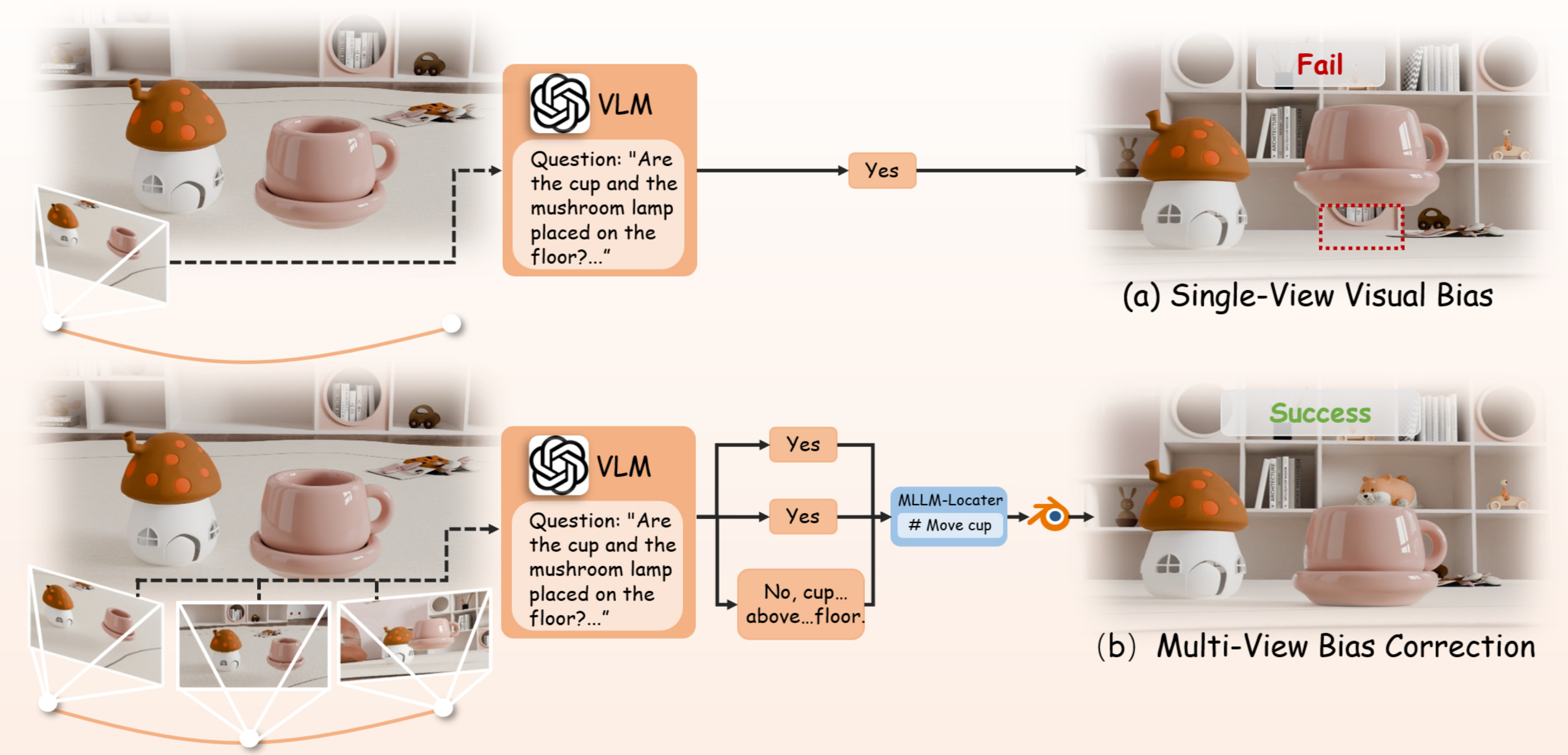}
    \vspace{-0.02cm}
    \caption{\textbf{Motivation for Multi-view Optimization. (a): }Optimizing the constructed scene from a single view may yield satisfactory results in that specific view, but object placements often exhibit semantic misalignments when observed from other perspectives. \textbf{(b): } Our proposed multi-view optimization effectively mitigates such inconsistencies by improving semantic correctness across multiple viewpoints, leading to globally coherent scene layouts.}
    \label{fig:multi-view}
\end{figure}
\noindent\textbf{Multi-View Optimization.} Despite constructing the whole scenes with the multimodal inputs, there still exists some mismatch in the whole layout. In the prior work~\cite{hu2024scenecraft}, they adopt an iterative visual feedback loop to refine scene layouts by leveraging an MLLM. However, such an approach based on a single rendered view is often insufficient, especially when dealing with physical interactions between objects. For instance, as shown in Fig.~\ref{fig:multi-view}, the input text condition is \textit{''Place two cups on a table.''}. If we only optimize the generated scene from a single viewpoint, it might only adjust the pink cup to a correct position in this view, while from another angle, the pink cup could still be floating above the table (as shown in Fig.~\ref{fig:multi-view} (a)). This discrepancy arises because the MLLM is unaware of depth inconsistencies hidden from the current view.

To ensure more reliable and physically grounded scene layouts, it is crucial to incorporate multi-view renderings during the feedback process. Multiple views enable the model to better verify spatial relationships, reducing visual illusions caused by limited viewpoints and producing more robust layouts. In our agent, we introduce a multi-view feedback optimization strategy, guided by a powerful Vision-Language Model (VLM). Our agent renders the current scene $L$ from multiple views $N = \{v_1, v_2, \dots, v_n\}$, and interacts with the VLM (e.g. \textit{''Is the pink cup placed on the table?''}) to verify the spatial relations for each view:  
\begin{equation}
    S_{\textit{{VLM}}} = \frac{1}{n} \sum_{i=1}^n s_i,
\end{equation}
where the $S$ is the confidence score feedback by VLM, if the scores exceed the threshold $t$, our agent determines that the current scene optimization is successful. Otherwise, the system creates a new location of object refinement via the \textit{MLLM-Locator}. The effectiveness of this strategy is illustrated in Fig.~\ref{fig:multi-view} (b), where a pink cup is initially misaligned from a side view, but after VLM-guided correction, it is accurately placed across all perspectives.

\begin{table*}[t!]
\begin{center} 
\resizebox{0.9\linewidth}{!}{
\begin{tabular}{l|c|cc|c}
\toprule
\multirow{3}{*}[0.8ex]{Method} & \multicolumn{1}{c|}{Scene Quality} & \multicolumn{2}{c|}{Scene Consistency} &\multicolumn{1}{c}{Text Align} \\
\cmidrule(lr){2-5} & Aesthetic $\uparrow$ & Subject Consis. $\uparrow$ & Background Consis. $\uparrow$  & CLIP Sim $\uparrow$ \\
\midrule
\multicolumn{5}{c}{ \textit{Proprietary MLLMs} }  \\ 
\midrule
Claude-4-Sonnet~\cite{Claude-4-Sonnet} & \textcolor{Blue}{\textbf{6.574}} & \textcolor{Blue}{\textbf{0.9135}} & 0.9078 & \textcolor{Blue}{\textbf{68.03}}\\
Claude-3.7-Sonnet~\cite{Claude-3.5-Sonnet} & 6.572 & 0.9122 & \textcolor{Blue}{\textbf{0.9079}} & 67.33 \\
Gemini-2.0-flash~\cite{Gemini2.0} & 6.472 & 0.9033 & 0.8925 & 65.48 \\
Gemini-2.5-Pro~\cite{comanici2025gemini} & 6.514 & 0.9138 & 0.9073 & 67.52 \\
Qwen-VL-Max~\cite{bai2023qwen} & 6.408 & 0.9131 & 0.9065 & 67.89 \\
GPT-4o~\cite{GPT-4o} & \textcolor{Red}{\textbf{6.592}} & \textcolor{Red}{\textbf{0.9144}} &  \textcolor{Red}{\textbf{0.9087}} & \textcolor{Red}{\textbf{68.75}} \\
GPT-4o-mini~\cite{GPT-4o-mini} & 6.137 & 0.9038 &  0.8977 & 63.49 \\
\midrule
\multicolumn{5}{c}{ \textit{Open-Source MLLMs} }  \\ 
\midrule
Llama-3.2-90B-Vision-Ins~\cite{llama3.2} & 6.357 & 0.9124 & 0.9057  & 66.35 \\
Llama-3.2-11B-Vision-Ins~\cite{llama3.2} & 6.324 &  0.9107 & 0.9031 & 65.17 \\
InternVL2.5-78B~\cite{chen2024expanding} & 6.209 & 0.9115 & 0.9062 & 65.23 \\
InternVL3-78B~\cite{zhu2025internvl3} & 6.213 & 0.9127 & 0.9075 & 67.18 \\
Qwen2.5-VL-72B-Ins~\cite{bai2025qwen2} & 6.312 & 0.9087 & 0.9011 & 65.47 \\
Qwen3-235B-A22B-Ins~\cite{yang2025qwen3technicalreport} & 6.379 & 0.9122 & 0.9043 & 66.12\\
Ovis2-34B~\cite{lu2024ovis} & 6.255 & 0.9081 & 0.9002 & 64.35 \\
Ovis2-16B~\cite{lu2024ovis} & 6.231 & 0.9075 & 0.8971 & 62.15 \\
gemma-3-27b-it~\cite{team2025gemma} & 6.204 & 0.9053 & 0.8953 & 60.35\\
gemma-3-12b-it~\cite{team2025gemma} & 6.197 & 0.9042 & 0.8891 & 59.67 \\

\bottomrule
\end{tabular}}
\caption{\textbf{Comparison with different MLLMs methods.} We generate 50 videos guided by each MLLM, and evaluate the performance of this data. The best result is in \textcolor{Red}{\textbf{Red}}, and the second best result is in \textcolor{Blue}{\textbf{Blue}}.}
\label{tab:data_3d} 
\end{center}
\end{table*}
\subsection{MLLM-Guided Task Planner}
\label{sec:frame prediction}

Even though the 2D images dataset generated by the \textit{MLLM-Optimizer} is sufficient for some simple tasks, such as 2D object removal, relighting, and inpainting. We aim to synthesize a high-quality video dataset for various practical applications. Equipped with the powerful ability of MLLM, e.g., in-context learning, long-term learning, we introduce the \textit{MLLM-Planner} for video generation. 

As shown in Fig.~\ref{fig:pipeline}, our \textit{MLLM-Planner} receives both the human instructions and the generated scenes as the input information. It first understands the visual scene and creates the visual state descriptions, locating the main object of the current frame. Then it reflects the human instructions and the feedback from the VLM-guided optimizer to refine the actions, facilitating to reasoning of the accuracy target objectives. Based on the reasoning results, it formulates the language plan and then converts it to an executable plan for generating the subsequence frames. 

However, a common issue remains that of temporal inconsistency across consecutive frames. This stems from the \textit{MLLM-Planner}, which focuses on discrete action execution without ensuring smooth transitions. As a result, generated sequences may suffer from abrupt changes, unnatural motion, or missing intermediate states. To address this, we introduce a VLM-guided frame prediction module (step 14 in Fig.~\ref{fig:pipeline}), which leverages VLM's visual reasoning to assess motion, object states, and scene dynamics across frames. Upon detecting inconsistencies, the module provides feedback to the \textit{MLLM-Planner}, prompting it to refine actions or insert intermediate steps. This iterative process enhances both temporal coherence and video quality.

\section{Experiments}
\begin{figure*}[t!]
    \centering
    \includegraphics[width=0.9\linewidth]{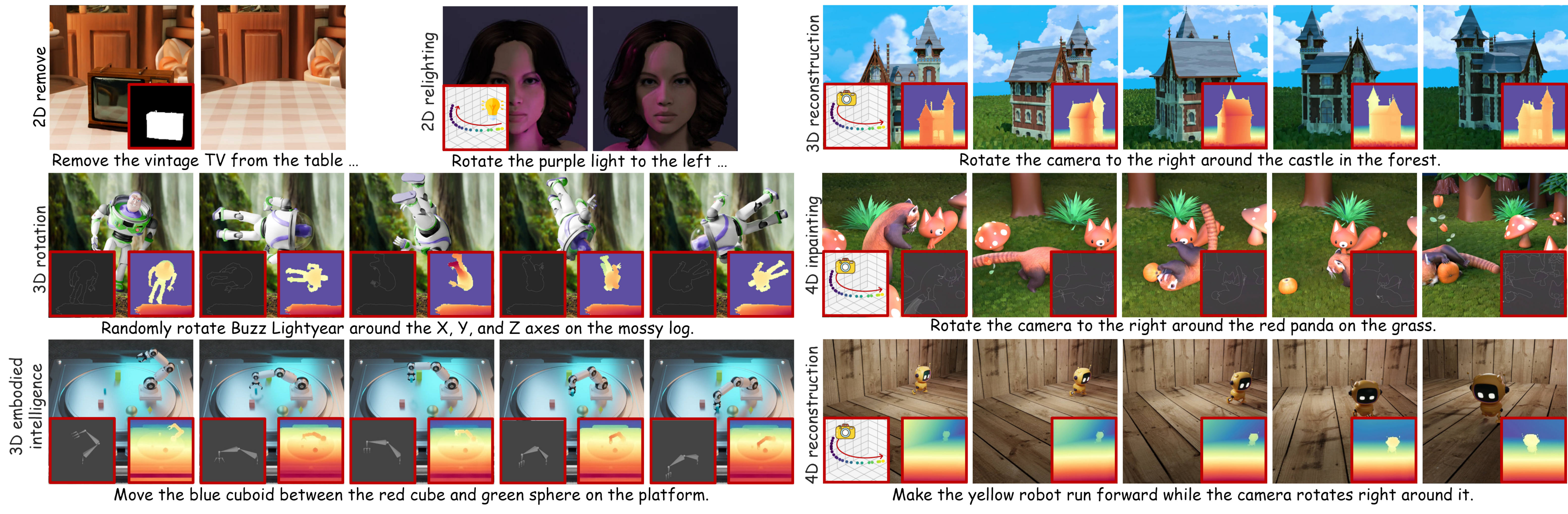}
    \caption{\textbf{Diverse downstream applications supported by Follow-Your-Instruction.} Each task is accompanied by tailored annotations, such as background masks for object removal, camera trajectories for relighting, 3D and 4D reconstruction, as well as depth maps and object poses for 3D embodied intelligence.
    } 
    \label{fig:applications}
\end{figure*}

\subsection{Evaluate the Quality of Generative Scenes}
\label{sec: evaluate data}
\textbf{Experimental Setup.} Most existing Multimodal Large Language Models (MLLMs) have shown promising capabilities in both vision and language understanding. Following the recent work~\cite{yangembodiedbench}, the primary baselines we compare against are state-of-the-art MLLMs, which can be categorized into closed-source proprietary models and open-sourced models, as they represent the current frontier of multimodal reasoning and decision-making.

Closed-source models include GPT-4o and GPT-4o-mini\cite{GPT-4o,GPT-4o-mini}, Claude-3.5-Sonnet and Claude-4-Sonnet~\cite{Claude-3.5-Sonnet, Claude-4-Sonnet}, Gemini-2.5-Pro, Gemini-2.0-flash~\cite{comanici2025gemini,Gemini2.0}, and Qwen-VL-Max~\cite{bai2023qwen}. These models are known for their strong performance in general multimodal tasks, with advanced reasoning abilities and extensive training on diverse internet-scale data. Open-sourced models, such as Llama-3.2 Vision Instruct~\cite{llama3.2}, InternVL2.5, InternVL3~\cite{chen2024expanding,zhu2025internvl3}, Qwen3, Qwen2.5-VL~\cite{yang2025qwen3technicalreport,bai2025qwen2}, Gemma-3~\cite{team2025gemma}, and Ovis2~\cite{lu2024ovis}, cover a range of model sizes (7B to 90B parameters) and provide accessible alternatives for research, allowing for deeper analysis of architectural design and scaling effects.

\noindent\textbf{Experimental Results.} Tab.~\ref{tab:data_3d} demonstrates a quantitative comparison of different MLLMs applied to our data-synthetic agents. We use the aesthetic score\cite{schuhmann2022laion} to assess perceptual quality, and measure scene consistency follows the VBench~\cite{huang2024vbench} in terms of subject appearance and background stability. Text alignment is evaluated via CLIP similarity~\cite{radford2021learning}. Results demonstrate the vital role of MLLM guidance in Follow-Your-Instruction, with GPT-4o achieving the best performance across all metrics, highlighting its superior cross-modal reasoning and alignment capabilities. Claude-4-Sonnet and Claude-3.7-Sonnet follow closely in aesthetics and consistency but lag in alignment. Among open-sourced models, InternVL3-78B and Qwen3-235B-A22B-Ins perform best overall, though a notable gap remains compared to GPT-4o. Noteably, this experiment is designed to highlight the generality of Follow-Your-Instruction’s core MLLM-driven capability across diverse AIGC tasks and MLLM structures, rather than focusing on achieving peak performance with any single MLLM. User study is provided in the supplementary materials.

\noindent\textbf{Applications.} As shown in Fig.~\ref{fig:applications}, we illustrate several representative tasks along with the corresponding ground-truth annotations generated by our agent. These examples highlight the agent's capacity to generalize across environments and task objectives. Our proposed agent's applications across a diverse set of tasks spanning 2D (object removal and re-lighting), 3D (reconstruction, rotation, and embodied intelligence), and 4D environments (4D inpainting and reconstruction).
These tasks reflect the \textbf{Follow-Your-Instruction's} potential for content creation in emerging research areas.

\begin{figure}[t]
    \centering
    \includegraphics[width=\linewidth]{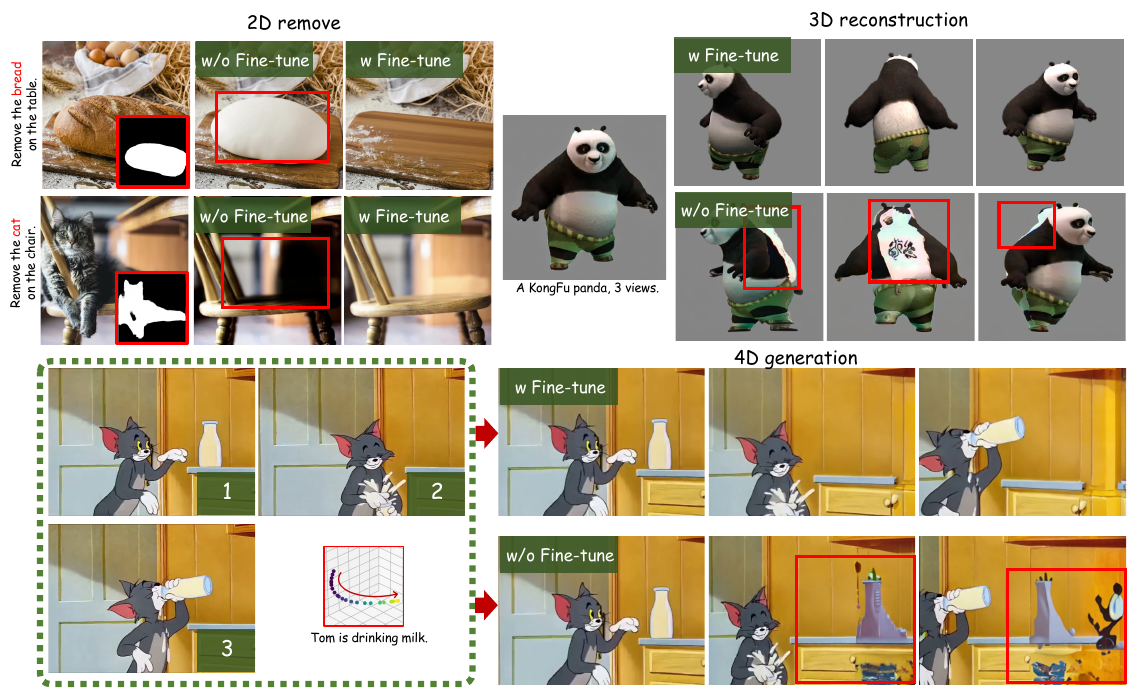}
    \caption{\textbf{Qualitative results for 2D object removal, 3D reconstruction, and 4D generation applications.} The results show that our generated data exhibits better effectiveness for improving the performance of existing models. 
    } 
    \label{fig:qualitative}
\end{figure}

\begin{table*}[t!]
	\begin{center}		
            \resizebox{\textwidth}{!}{
    		\begin{tabular}{l|cccc|cc|ccc}
    			\toprule
    			\multirow{3}{*}{Method} & \multicolumn{4}{c|}{Visual Quality} & \multicolumn{2}{c|}{Camera Accuracy} & \multicolumn{3}{c}{View Synchronization}\\ 
                    \cmidrule(r){2-10}
                    & \makecell[c]{FID $\downarrow$} & \makecell[c]{FVD $\downarrow$} & \makecell[c]{CLIP-T $\uparrow$} & \makecell[c]{CLIP-F $\uparrow$} & RotErr $\downarrow$& TransErr $\downarrow$ & {Mat. Pix.(K) $\uparrow$} & {FVD-V $\downarrow$} & {CLIP-V $\uparrow$}\\
                    \midrule 
                    
                    ReCamMaster~\cite{bai2025recammaster} & 62.48 & 160.72 & 34.97  & 96.23  & 1.45  & 5.22  & 630.51 & 151.28 & 88.59 \\
                    \midrule
                    + Our data & \textbf{60.32} & \textbf{155.71} & \textbf{35.88}  & \textbf{96.66} & \textbf{1.35} & \textbf{4.69} & \textbf{682.57} & \textbf{135.24} & \textbf{89.92}  \\
    			\bottomrule
    		\end{tabular}
            \label{tab:othermetrix}
            }
            \caption{\textbf{Quantitative results for 4D generation tasks.} We assess visual quality, camera accuracy, and view synchronization. The best results are in bold. }
            \label{tab:com_4d}  %
	\end{center}
\end{table*}

\subsection{Evaluation on Downstream Application}
\label{sec: evaluate model}
\noindent\textbf{Baselines.} To further evaluate the quality of the synthetic data, we fine-tune several baseline models on both 2D/3D/4D AIGC applications, including object removal, 3D reconstruction, and 4D video generation.  For the 2D object removal task, we adopt RoRem~\cite{li2025rorem} as the baseline and assess the improvement after fine-tuning with our data. For the 3D reconstruction task, we utilize MV-Adapter~\cite{huang2024mv}, a recent multi-view reconstruction framework, and evaluate performance improvements in terms of geometry accuracy and consistency. For the 4D video generation task, we employ ReCamMaster~\cite{bai2025recammaster} to measure temporal coherence and fidelity in dynamic scene synthesis. These baselines allow us to systematically quantify the impact of our synthetic data on different aspects of AIGC models across various dimensions.

\noindent\textbf{Qualitative Results.} The visual comparison for the 2D, 3D, and 4D applications is shown in Fig.~\ref{fig:qualitative}. We can see that before fine-tuning with our generated data, the performance of object removal is struggled with the semantic completion (as shown in the first row for the 2D task in Fig.~\ref{fig:qualitative}, the model generate a strange white object rather than inpaint the cutting board), and generates some artifacts after removing (the second row for the 2D task in Fig.~\ref{fig:qualitative}). In contrast, after fine-tuning with our generated data, these problems have been alleviated significantly. On the other hand, for the 3D task, without fine-tuning with our data, although the model can generate front views well, it struggles to generate high-quality and consistent back views, and the hallucination issue can be fixed after the fine-tuning process. 

Additionally, the 4D generation task is an emerging paradigm of controllable video synthesis under the guidance of camera trajectories. As shown in Fig.~\ref{fig:qualitative}, while the ReCamMaster achieves better pose accuracy and smooth camera movement, there are still some inconsistencies and artifacts in the background, and our generated data has improved its performance.

\noindent\textbf{Quantitative Results.} We also perform the quantitative experiments for the three applications. The quantitative results of 2D object removal and 3D reconstruction can be found in the supplementary material, and the results of 4D generation are shown in Tab.~\ref{tab:com_4d}. Following the ReCamMaster~\cite{bai2025recammaster}, we evaluate the visual quality, camera accuracy, and the view synchronization. Specifically, we calculate the rotation and translation errors to evaluate the camera trajectory accuracy, and compute the CLIP-V and FVD-V to assess the view synchronization between the different viewpoints in the same scene. The results show that the performance of the baseline model can be improved after fine-tuning.

\subsection{Ablation Study}
\label{sec: ablation}

\begin{figure}[t!]
    \centering
    \includegraphics[width=0.8\linewidth]{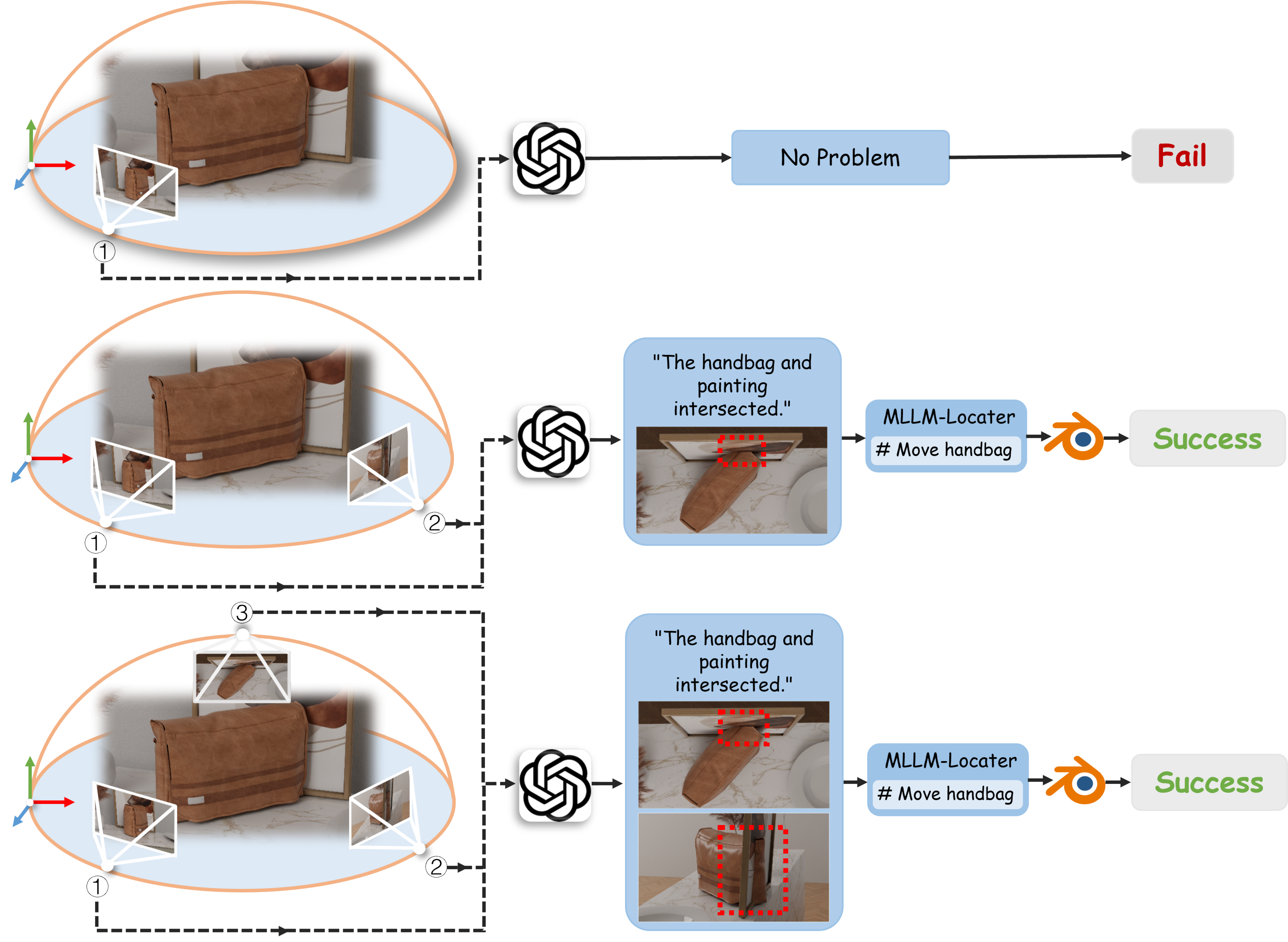}
    \caption{\textbf{Ablation study about the Multi-view optimization.} The global scene still has misalignment, only optimizing with a single view. (first row). And this issue can be addressed by increasing the number of reference views (second and third rows). }
    \label{fig:ablation_multi}
\end{figure}

\noindent\textbf{Effectiveness of the Multi-view optimization.} As shown in Fig.~\ref{fig:ablation_multi}, we evaluate the performance of different numbers of frames in the Multi-view optimization strategy. When we use only one view to optimize the generated scene, although in the current view the object has been located in the correct position, it often fails to align properly in other unseen perspectives. This limitation can be mitigated by incorporating more views during optimization. Furthermore, we also conduct a quantitative ablation study to assess the proper number of views we need to provide in this process, and the results are shown in the Tab.~\ref{tab:ablation_multiview}. We can observe that increasing the number of views leads to longer generation times, while offering only marginal improvements in optimization success rate. Based on this analysis, we select two views as the optimal setting, balancing both efficiency and performance.

\begin{table}[h!]
\centering

\begin{tabular}{c|c|c|c}
\toprule
{Num of Views} & \multicolumn{1}{c|}{CLIP Sim $\uparrow$} & \multicolumn{1}{c|}{Success Rate $\uparrow$} &\multicolumn{1}{c}{Time (s)} \\

\midrule
1 & 53.24 & 0.2415 & \textbf{380} \\
2 & \textbf{68.75} & 0.9987 & 386 \\
3 & 68.63 & \textbf{0.9994} & 398 \\
\bottomrule
\end{tabular}
\caption{\textbf{Quantitative ablation results for Multi-view optimization (step 7 in Fig.~\ref{fig:pipeline}).} The results demonstrate that with the increase of views, the quality of scenes exhibits a trend of first improving and then keeping the same level, while an increase in time has raised the cost. A view count of 2 offers the best balance.}
\label{tab:ablation_multiview}  %
\end{table}

\begin{table}[h!]
\centering

\begin{tabular}{c|c|c}
\toprule
{Method} & \multicolumn{1}{c|}{CLIP Sim $\uparrow$} & \multicolumn{1}{c}{Temporal Consis. $\uparrow$} \\

\midrule
w/o VLM & 50.76 & 0.6524  \\
ours & \textbf{68.75} & \textbf{0.9128}  \\

\bottomrule
\end{tabular}
\caption{\textbf{Ablation results for VLM-guided frame prediction (step 14 in Fig.~\ref{fig:pipeline}).} The best results are in bold.}
\label{tab:ablation_vlm}  %
\end{table}
\begin{figure}[t!]
    \centering
    \includegraphics[width=0.95\linewidth]{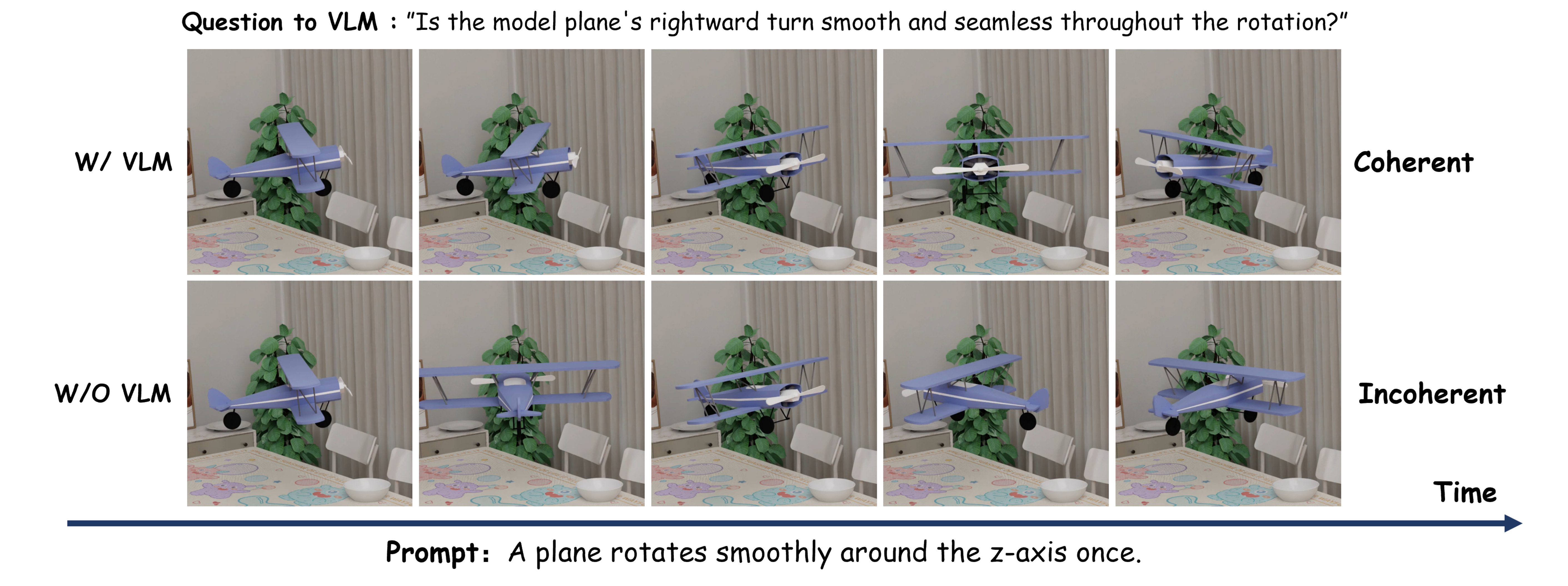}
    \caption{\textbf{Ablation study about the VLM-guided frame prediction.} The second row shows that without the VLM optimization, the plane exhibits abrupt, unsmooth motion with rapid successive turns.}
    \label{fig:ablation_vlm}
\end{figure}
\noindent\textbf{Effectiveness of VLM-guided frame prediction.} We also evaluate the contribution of the VLM-guided frame prediction in Fig.~\ref{fig:ablation_vlm} and Tab.~\ref{tab:ablation_vlm}. In the second row of Fig.~\ref{fig:ablation_vlm}, it can be observed that without the VLM-guided frame prediction strategy, the resulting video often suffers from temporal inconsistency, where the motion between adjacent frames is abrupt and unsmooth. Specifically, the rotation angle of the airplane is excessively large between two consecutive frames, making it appear as if the plane turns twice within a short duration. This indicates that the planned actions lack continuity, resulting in suboptimal visual quality and temporal inconsistency.

\section{Conclusion and Discussion}
\textbf{Conclusion.} This paper has proposed the \textbf{Follow-Your-Instruction}, an efficient MLLM-based data-synthetic agent framework, which generates the realistic scenes across 2D, 3D, and 4D levels from the multimodal inputs (e.g., text, image, or the blend file). \textbf{Follow-Your-Instruction} is built on the Multimodal Large Language Models, combining with four main components: \textit{MLLM-Collector}, \textit{MLLM-Generator}, \textit{MLLM-Optimizer}, and \textit{MLLM-Planner}. First, the \textit{MLLM-Collector} transforms the text input into the assets or integrates the assets from the visual inputs, enhancing the user-oriented scene creation. Then the \textit{MLLM-Generator} creates the 3D layout for the scene and optimizes it via \textit{MLLM-Optimizer}. Finally, the \textit{MLLM-Planner} creates the subsequence frames and refines them by the VLM-guided frame prediction module. Experimental results demonstrate that our agent leverages the capability of the MLLMs in our data synthesis process, facilitating several downstream AIGC applications.  

\noindent\textbf{Limitations.} There are three limitations of our method: (1) the performance relies on the capabilities of its underlying proprietary MLLM; (2) the lack of validation of generated data in improving generalization to other real-world benchmarks; (3) the scalability is constrained by its reliance on pre-existing asset libraries. 
{
    \small
    \bibliographystyle{ieeenat_fullname}
    \bibliography{main}
}

\end{document}